%% file: root.tex
\documentclass[letterpaper, 10 pt, conference]{ieeeconf}  %

\IEEEoverridecommandlockouts                              %

\makeatletter
\let\NAT@parse\undefined
\makeatother

\usepackage{graphics,mathptmx,times,amssymb}
\usepackage{amsmath,epsfig,xcolor,graphicx,url,xspace,booktabs}
\usepackage{url,multirow,bm,bbm,breqn,rotating,color,colortbl,subcaption}

\let\labelindent\relax
\usepackage{enumitem}

\definecolor{Gray}{gray}{0.86}

\usepackage{hyperref}
\definecolor{citecolor}{HTML}{0071BC}
\definecolor{linkcolor}{HTML}{ED1C24}

\hypersetup{colorlinks=True, citecolor=citecolor, urlcolor=magenta}

\title{\LARGE \bf
Barely-Visible Surface Crack Detection\\for Wind Turbine Sustainability
}

\author{Sourav Agrawal, Isaac Corley, Conor Wallace, Clovis Vaughn, and Jonathan Lwowski%
\thanks{All authors are with the AI Research team at Zeitview
        {\tt\small \{sourav.agrawal, isaac.corley, conor.wallace, clovis.vaughn, jonathan.lwowski\}@zeitview.com}}%
}

\begin{document}
\makeatletter
\g@addto@macro\@maketitle{
  \captionsetup{type=figure}\setcounter{figure}{0}
  \def\mycolspace{1.2mm}
  \centering
  \vspace{1ex}
    \includegraphics[width=1.0\textwidth]{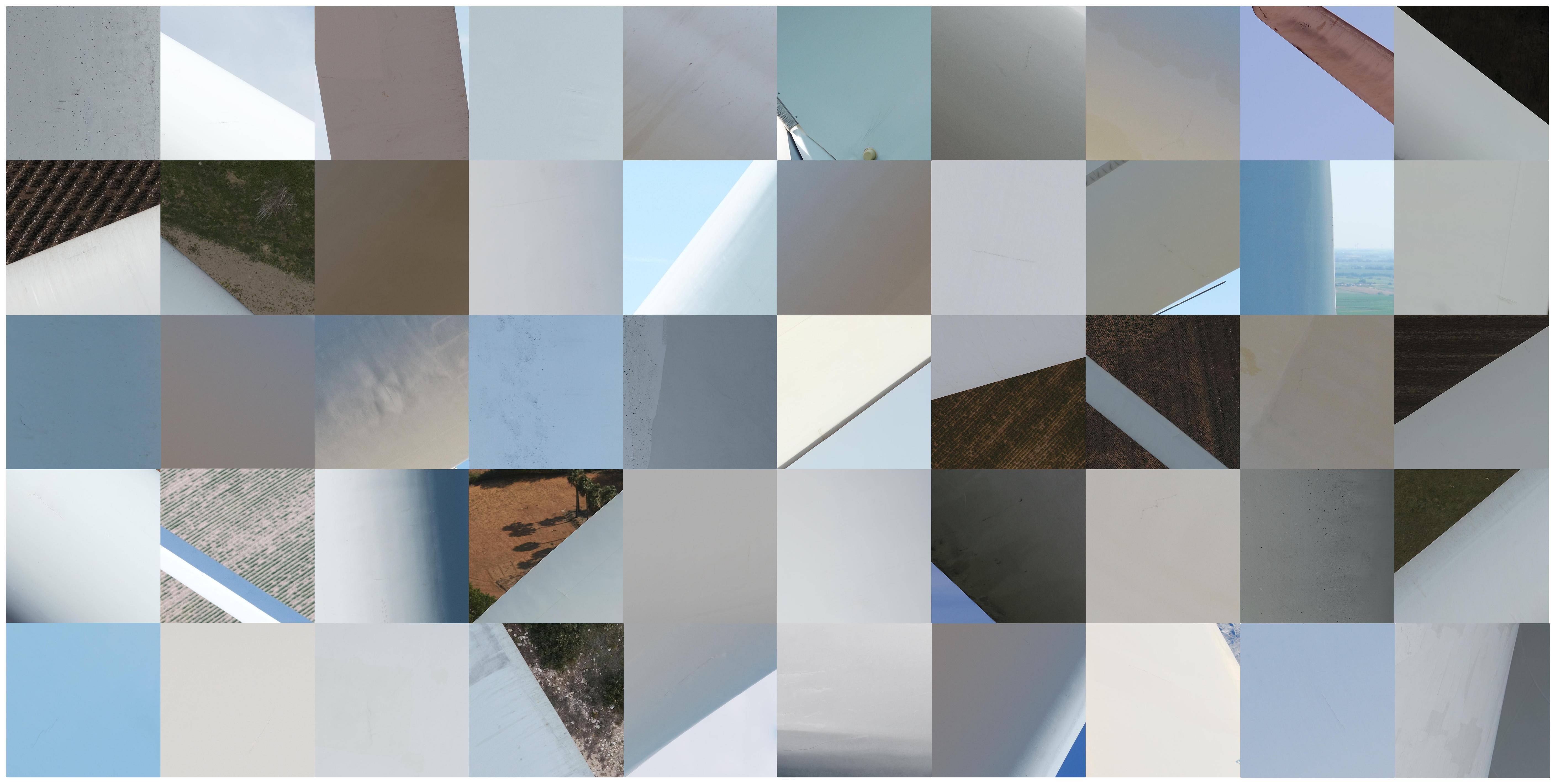}
	\caption{\textbf{Can you spot the cracks in these images?} Hairline cracks are barely visible and commonly confused with stains and markings from dirt or grease. However, these defects can lead to catastrophic and irreparable damage to wind turbines if left untreated. This figure contains sample tiles from our \texttt{Zeitview Crack Detection (ZVCD)} dataset of annotated barely-visible surface cracks collected from inspections across the globe. This dataset is used to train our turbine crack detection pipeline for the inspection of renewable wind energy sources.}
	\label{fig:samples}
}
\makeatother
\maketitle
\thispagestyle{empty}
\pagestyle{empty}

\begin{abstract}

The production of wind energy is a crucial part of sustainable development and reducing the reliance on fossil fuels. Maintaining the integrity of wind turbines to produce this energy is a costly and time-consuming task requiring repeated inspection and maintenance. While autonomous drones have proven to make this process more efficient, the algorithms for detecting anomalies to prevent catastrophic damage to turbine blades have fallen behind due to some dangerous defects, such as hairline cracks, being barely-visible. Existing datasets and literature are lacking and tend towards detecting obvious and visible defects in addition to not being geographically diverse. In this paper we introduce a novel and diverse dataset of barely-visible hairline cracks collected from numerous wind turbine inspections. To prove the efficacy of our dataset, we detail our end-to-end deployed turbine crack detection pipeline from the image acquisition stage to the use of predictions in providing automated maintenance recommendations to extend the life and efficiency of wind turbines.
\end{abstract}

\section{INTRODUCTION}
\input{Sections/intro.tex}
\label{sec:intro}

\section{METHODS}
\input{Sections/methods.tex}
\label{sec:methods}

\section{EXPERIMENTS}
\input{Sections/experiments.tex}
\label{sec:experiments}

\section{PIPELINE}
\input{Sections/pipeline.tex}

\label{sec:pipeline}

\label{sec:future-work}

\input{Sections/future_work.tex}
\section{CONCLUSION}
\label{sec:conclusion}
\input{Sections/conclusion.tex}

\addtolength{\textheight}{-9cm}   %

\bibliographystyle{IEEEtran}
\bibliography{refs}

\end{document}

%% file: Sections/intro.tex
Wind energy stands as a pivotal component in the pursuit of sustainable development, playing a crucial role in mitigating the impacts of climate change and reducing reliance on fossil fuels. However, the efficient and long-term operation of wind turbines, which are at the heart of wind energy production, is encumbered by the challenges of maintaining their structural integrity. Traditional methods of turbine inspection are not only time-consuming and costly but also fall short in detecting minute yet critical anomalies, such as barely-visible surface cracks, that can lead to irreparable damage and catastrophic failures costing upwards of \$5-8M to replace a modern 3.3MW onshore wind turbine~\cite{Weather_Guard_Lightning_Tech_2023,Hartman_2023}. The advent of autonomous drones has revolutionized this process, offering a more efficient means of monitoring the condition of turbine blades. Nevertheless, the true potential of this technology is hindered by the limitations of current machine learning models, such as object detection and segmentation, often requiring extensive post-processing to eliminate erroneous predictions when applied in real-world scenarios.

\subsection{Background}
\label{sec:background}

\hspace{-2ex}\textbf{Surface Crack Types}\hspace{1ex} Among the different classifications of damages wind turbine blades experience, structural cracks are one of the most critical yet also generally undetectable with routine external inspections performed by humans~\cite{mishnaevsky2022root}. Blade cracks can originate in many ways and their impact to the blade structure may vary. Coating cracks may occur due to impact from foreign objects, fatigue, or secondary to an underlying structural defect such as delamination. Laminate cracks can also occur from prolonged fatigue and overload events, or they can form over time due to manufacturing defects such as wrinkles in the laminate layers or gaps in the core materials~\cite{dashtkar2019rain}.

Once a crack forms, it is difficult to predict how quickly it will grow. Failure rates can occur within hours of detection to years depending on size and location. The structural significance to the blade is difficult to determine without referencing the unique specifications of each particular blade model, which is not always possible~\cite{katsaprakakis2021comprehensive}. As a result, it is important to identify and classify cracks as soon as possible so that the appropriate repairs can be made and further damage can be mitigated.

In most cases, cracks will appear as black lines on a blade that is painted white, but the length and width of the line can vary significantly. These cracks may only be a few centimeters long and less than a millimeter wide in some cases, making identification difficult even on a clean blade that is fully painted white~\cite{wang2022review}. Furthermore, detection can suffer when the blade is dirty or when image quality is poor. One of the greatest sources of a false identification is a dirt or grease mark that has streaked across the blade. There are subtle visual differences between these cases and true cracks, resulting in automated methods which struggle with overproducing false positives.

\vspace{1ex}\hspace{-2ex}\textbf{Severity Categorization}\hspace{1ex} There is no universal industry standard for blade defect categorization, however, there is a commonly accepted \textit{best practice} which classifies damages into one of five ordinal severity levels~\cite{Malkin_2020b}, ranging from low to high risk of existing or potential future damage to the blade structure. Cosmetic defects and surface-level damage generally fall within the \texttt{Severity [1-2]} range. \texttt{Severity 3} can encompass larger surface-level defects, or early, low risk structural damage. \texttt{Severity [4-5]} defects are generally considered to be moderate to high severity structural damage. Cracks typically fall within the \texttt{Severity [3-5]} range, with emerging cracks, referred to as \textit{hairline cracks}, often falling under the lower end of that range~\cite{epri_2020}.

\vspace{1ex}\hspace{-2ex}\textbf{Related Work}\hspace{1ex} While there are numerous prior works in the field of external anomaly and defect detection in wind turbine blades~\cite{gohar2023automatic,gohar2023slice,zhang2022image}, few of them focus specifically on crack detection~\cite{wang2017automatic,wang2019two,xiaoxun2022research,joshuva2019crack}. Other works utilize infrared cameras for detecting cracks which can provide insights not apparent in the visible sprectrum~\cite{jia2018rapid, jaeger2022infrared, wang2022research}. However the hardware to effectively make use of these thermography-based algorithms are more costly than RGB cameras, hence, their slow adoption in industry~\cite{nguyen2021review}.

However, all of these works focus specifically on easily visible and detectable surface cracks. Our work focuses specifically on barely visible and high severity hairline cracks which can easily go undetected until it is too late and catastrophic damage occurs to the wind turbine between inspections that is beyond repair.

\subsection{Contributions}

In this work, we address the aforementioned challenges by introducing a comprehensive approach that combines the capabilities of autonomous drones with machine learning specifically for detecting barely-visible cracks in wind turbine blades. Our approach involves all aspects of machine learning from dataset curation to model training, resulting in an accurate and end-to-end intelligent system for automated inspection of wind turbine blades. We detail a fully-deployed pipeline for more reliable and efficient maintenance of wind turbines, ultimately contributing to the sustainable production of wind energy. Our contributions are as follows:

\vspace{1ex}
\begin{description}
    \item[Zeitview Crack Detection (\texttt{ZVCD}) Dataset] Throughout several years of inspections and numerous turbines analyzed for defects, we have compiled a novel, large, high severity surface crack detection dataset. We find a dataset of this size, geographical diversity, and quality key in producing machine learning models which can accurately detect barely-visible turbine blade cracks.
    
    \vspace{1ex}
    \item[Deployed Turbine Crack Detection Pipeline] We detail Zeitview's end-to-end crack detection pipeline, providing industry insights for how to deploy a model trained on our \texttt{ZVCD} dataset at scale in real-world wind turbine inspection scenarios.
\end{description}

\input{tables/datasets}

%% file: tables/datasets.tex
\begin{table*}[ht!]
\centering
\resizebox{0.98\linewidth}{!}{%
\begin{tabular}{lcccccccc}
\toprule
\textbf{Dataset} &
\textbf{\# Cracks} &
\textbf{Barely Visible} &
\textbf{High Severity} &
\textbf{Real} &
\textbf{\# Locations} &
\textbf{\# Turbines} &
\textbf{\# Manufacturers} &
\textbf{\# Models} \\
\toprule
Blade30~\cite{yang2023towards} & 9 & $\times$ & $\times$ & \checkmark & 6 & - & - & - \\
IDR~\cite{yu2017image} & 12 & $\times$ & $\times$ & \checkmark & 1 & - & - & - \\
UAD~\cite{wang2019unsupervised} & 30 & $\times$ & $\times$ & \checkmark & 1 & - & - & - \\
ADO~\cite{wang2017automatic} & 50 & $\times$ & $\times$ & \checkmark & 1 & 33 & - & - \\
IEM~\cite{zhang2021image} & 55 & $\times$ & $\times$ & \checkmark & 1 & - & - & - \\
IRO~\cite{yang2021image} & $<$200 & $\times$ & $\times$ & \checkmark & 1 & - & - & - \\
CAI-SWTB~\cite{altice2024anomaly} & 258 & $\times$ & $\times$ & $\times$ & 1 & - & - & - \\
DOC~\cite{reddy2019detection} & 2,189 & $\times$ & $\times$ & \checkmark & 1 & - & - & - \\
\rowcolor{Gray}
\textbf{\texttt{ZVCD} (Ours)} & 9,107 & \checkmark  & \checkmark & \checkmark & 988 & 4,684 & 34 & 195 \\
\bottomrule
\end{tabular}%
}
\caption{\textbf{Comparison of turbine blade crack detection datasets.} Existing datasets suffer from several issues such as being simulated or using image augmentations to artificially inflate the dataset size. Further, these datasets typically contain visibly obvious and/or low severity defects by industry standards, creating them infeasible for detecting barely-visible surface cracks. Our proposed \texttt{ZVCD} dataset consists of real inspection imagery from a diverse set of geographic sites and turbine models and manufacturers. Note that some literature does not disclose how many unique turbines, model, and manufacturers are captured in their datasets. Some datasets use image augmentations to artificially inflate number of images in the dataset, therefore we report number of unique cracks in each dataset.}
\label{tab:datasets}
\end{table*}

%% file: Sections/methods.tex
\subsection{Dataset}
\label{sec:datasets}

There is a scarcity of existing turbine anomaly detection datasets. Many of the recently proposed datasets are proprietary making them difficult to benchmark against. Additionally, several prior works suffer from issues such as only containing simulated imagery or using intensive image augmentations to artificially inflate the dataset size. Furthermore, of the few datasets that are available, crack detection is only a subset of the anomaly categories making it a relatively rare defect, further limiting the amount of available data to train and evaluate against. Rarer still are hairline cracks which are barely visible and only occur on a small subset of turbines. This makes it incredibly difficult to collect a comprehensive dataset. For example, after inspecting roughly 35,000 turbines, we expect to observe approximately 3,000 turbines with hairline cracks.

\vspace{1ex}\hspace{-2ex}\textbf{ZVCD dataset}\hspace{1ex} Hairline cracks have a wide variety of appearances ranging from visible and great in length to nearly invisible and short in length. Wind turbines can also vary in model type and manufacturer as well as location across the globe. Machine learning models trained to detect defects need to generalize to these variations and be robust enough to learn to extract features only on the blade. Therefore, we create a comprehensive, geographically diverse dataset of barely-visible cracks from wind turbine inspections which we refer to as the \texttt{Zeitview Crack Detection (ZVCD)} dataset. Figure \ref{fig:locations} displays a map of the turbine farm locations captured in our dataset. Table \ref{tab:datasets} describes the attributes of our dataset compared to pre-existing datasets found in the literature. We observe a much greater diversity of wind turbines and locations, thus allowing our models to generalize better when deployed in a real-world inspection scenario.

\subsection{Models}
\label{sec:models}

\hspace{-2ex}\textbf{Crack Detection Modeling in Practice}\hspace{1ex}  While segmentation and object detection methods are popular for detecting turbine defects and potentially provide more accurate localization, we instead choose to utilize classification for this task for the following reasons:

\begin{itemize}
    \item Turbine inspection imagery are commonly acquired at high resolutions resulting in images too large to use with modern machine learning models, therefore images are patched into sub-tiles making localization of detections not as critical because of the known location of the sub-tiles.
    \item Segmentation and Object Detection are sensitive and commonly require significant post-processing to remove erroneous predictions and false positives.
    \item Segmentation and object detection labels are more costly to annotate in comparison to classification.
\end{itemize}

While classification isn't as appealing as the alternatives, we find in practice that crack detection classifiers operating on sub-tiles of an image dramatically reduce the amount of false positive detections while also maintaining efficiency for use with on-board hardware or deployed on a server.

\vspace{1ex}\hspace{-2ex}\textbf{Model Architectures}\hspace{1ex}  In our experiments we train the following classifier models on our \texttt{ZVCD} dataset. We select several models ranging in parameter sizes while targeting efficiency for use on-board drone hardware. All models are initialized with ImageNet pretrained weights~\cite{deng2009imagenet}.

\input{figures/locations}
\input{tables/results}

\begin{itemize}
    \item \textbf{ResNet-18}:~\cite{he2016deep} is a canonical ConvNet classifier model which uses skip connections after each block to improve flow of features and gradients.
    \item \textbf{EfficientNet}~\cite{tan2020efficientnet} is a model designed specifically to maintain similar performance to large classifier models while reducing the number of parameters to improve efficiency and usage in hardware constrained settings. In our experiments we utilize the EfficientNet-B3 variant.
    \item \textbf{MobileNetV3}~\cite{howard2019searching} is a model designed using neural architecture search (NAS) methods to both improve performance on mobile hardware while maintaining performance on ImageNet. In our experiments we utilize the MobileNetV3 Large and Small variants.
\end{itemize}

%% file: figures/locations.tex
\begin{figure}[t!]
\centering
\includegraphics[width=0.98\linewidth]{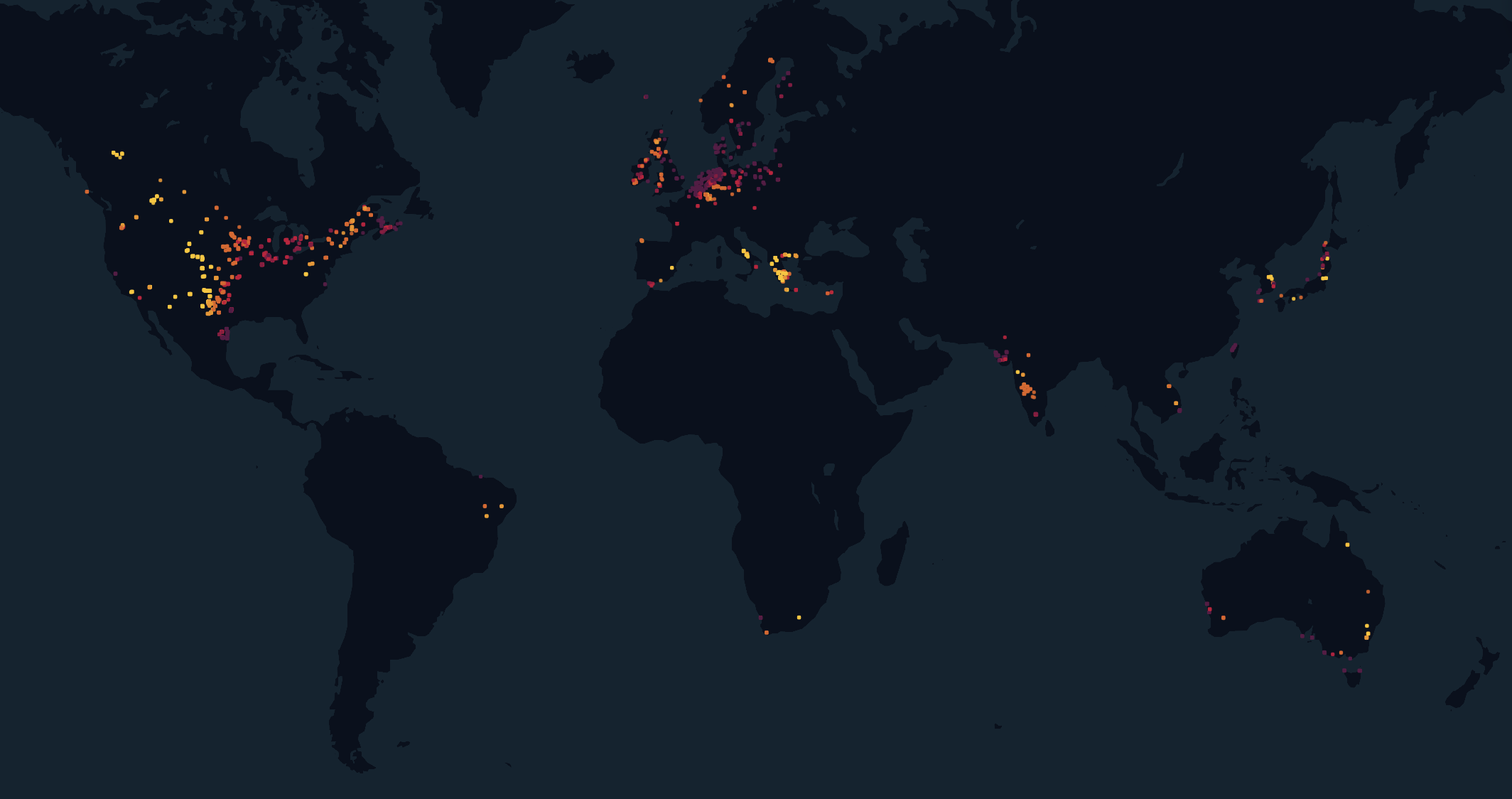}
\caption{\textbf{Geographic locations of the inspections in our \texttt{ZVCD} dataset.} Our dataset consists of a diverse set of imagery of barely-visible surface cracks acquired from 4,684 turbine inspections from nearly 1,000 unique sites across the world.}
\label{fig:locations}
\end{figure}

%% file: tables/results.tex
\begin{table*}[ht!]
\centering
\resizebox{0.98\linewidth}{!}{%
\begin{tabular}{ccccccccc}
\toprule
\textbf{Backbone} &
\textbf{\# Params (M)} &
\textbf{FLOPs (G)} &
\textbf{Runtime (ms)} &
\textbf{ImageNet Acc.} &
\textbf{Accuracy} &
\textbf{F1} &
\textbf{Precision} &
\textbf{Recall} \\
\toprule
ResNet-18~\cite{he2016deep} & 11.2 & 38.00 & 15 & 73.16 & 70.34 & 66.94 & 63.85 & 70.34 \\
EfficientNet-B3~\cite{tan2020efficientnet} & 10.7  & 20.59 & 47 & \textbf{80.87} & \textbf{82.20} & 74.33 & 67.83 & \textbf{82.20} \\
MobileNetV3-S~\cite{howard2019searching} & 1.5  & 1.20 & 11 & 67.92 & 81.35 & \textbf{74.42} & 68.57 & 81.36 \\
MobileNetV3-L~\cite{howard2019searching} & 4.2  & 4.63 & 16 & 75.512 & 74.58 & 72.73 & \textbf{70.97} & 74.58 \\

\bottomrule
\end{tabular}%
}
\caption{\textbf{Test set performance of classifier models trained on our \texttt{ZVCD} dataset.} We report accuracy, f1-score, precision, and recall. Because missing cracks can be costly, we place more emphasis on precision when determining the best performing model. We further report metrics for model size in millions (M) of parameters and runtime efficiency in milliseconds (ms).}
\label{tab:results}
\end{table*}

%% file: Sections/experiments.tex
 \subsection{Preprocessing}
To generate our datasets for training and evaluation we preprocess the \texttt{ZVCD} dataset described in Section~\ref{sec:datasets}. Our dataset contains roughly annotated polygons and a severity category for each crack labeled by our inspection analysts. We then tile each image into patches with a 25\% overlap, keeping all patches that contain a crack and 1\% of patches which do not contain a crack. We then filter by severity, removing all \texttt{Severity [1-2]} cracks, and further filter easily visible cracks.

\subsection{Dataset Splits}
Given our preprocessed dataset, we then split our dataset by turbine, not by image, using a 90/10 train/val split criterion, such that no tile from the same image will be in both sets. We further manually create a very hard subset of our val set as test set which contains the least visible surface cracks. Samples of this set can be viewed in Figure~\ref{fig:samples}.

\subsection{Metrics}
For each model variant, we report the original ImageNet accuracy from pretraining for clarity. Additionally we report the accuracy, F1 score, precision, and recall on the \texttt{ZVCD} very hard test set. We place particular importance on precision as our objective is to minimize missed crack detections.

\subsection{Training Details}
To keep comparisons fair, we retain the same training details across all model variants. We use an initial learning rate of $10^{-3}$ paired with the AdamW optimizer~\cite{loshchilov2017decoupled} and a cosine learning rate scheduler. The dataset is tiled with an image size of 1024 $\times$ 1024 with an overlap of 25\%. We train for 50 epochs with no early stopping condition. For data loading, we use a batch size of 8 images and we apply a balanced weighting to the two classes to eliminate imbalance bias towards the negative class.

We use the model architectures implemented in the PyTorch Image Models (timm) library~\cite{rw2019timm} and trained using the PyTorch Lightning library~\cite{pytorch-lightning}. All models were trained with the same train-time image augmentations: 1) random horizontal flip with $p=0.5$, 2) random vertical flip with $p=0.5$, 3) random affine transform with angle between $\alpha=[0,90]$ and $p=0.5$, and 4) color jitter with $brightness=0.1$, $contrast=0.1$, $saturation=0.1$, $hue=0.1$ and $p=0.25$. We use implementation from the Kornia library~\cite{riba2020kornia}. All images are preprocessed by normalizing with ImageNet statistics of $mean=[0.485, 0.456, 0.406]$ and a $std=[0.229, 0.224, 0.225]$. All experiments were run on an AWS compute instance with 8 vCPUs and a Tesla V100-SXM2-16GB GPU.

\input{figures/val_metrics}

\subsection{Training Results}
We report the metrics for each model on the \texttt{ZVCD} test set in Table~\ref{tab:results}. For additional insight we display the validation accuracy performance throughout training in Figure~\ref{fig:val_metrics}. We further compute efficiency metrics -- total number of parameters, floating-point operations per second (FLOPS), and runtime (ms) -- for each model using the fvcore~\cite{fvcore} library with a batch size of 1 and an image size of $3 \times 1024 \times 1024$.

While these models are by no means bleeding edge, we found that popular state-of-the-art transformer-based architectures like ConvNextV2~\cite{woo2023convnext} and SwinV2~\cite{liu2022swin} would not converge to the same performance during training, likely requiring significant hyperparameter tuning to reach similar performance to the other models we used.

%% file: figures/val_metrics.tex
\begin{figure}[t!]
\centering
\includegraphics[width=0.8\linewidth]{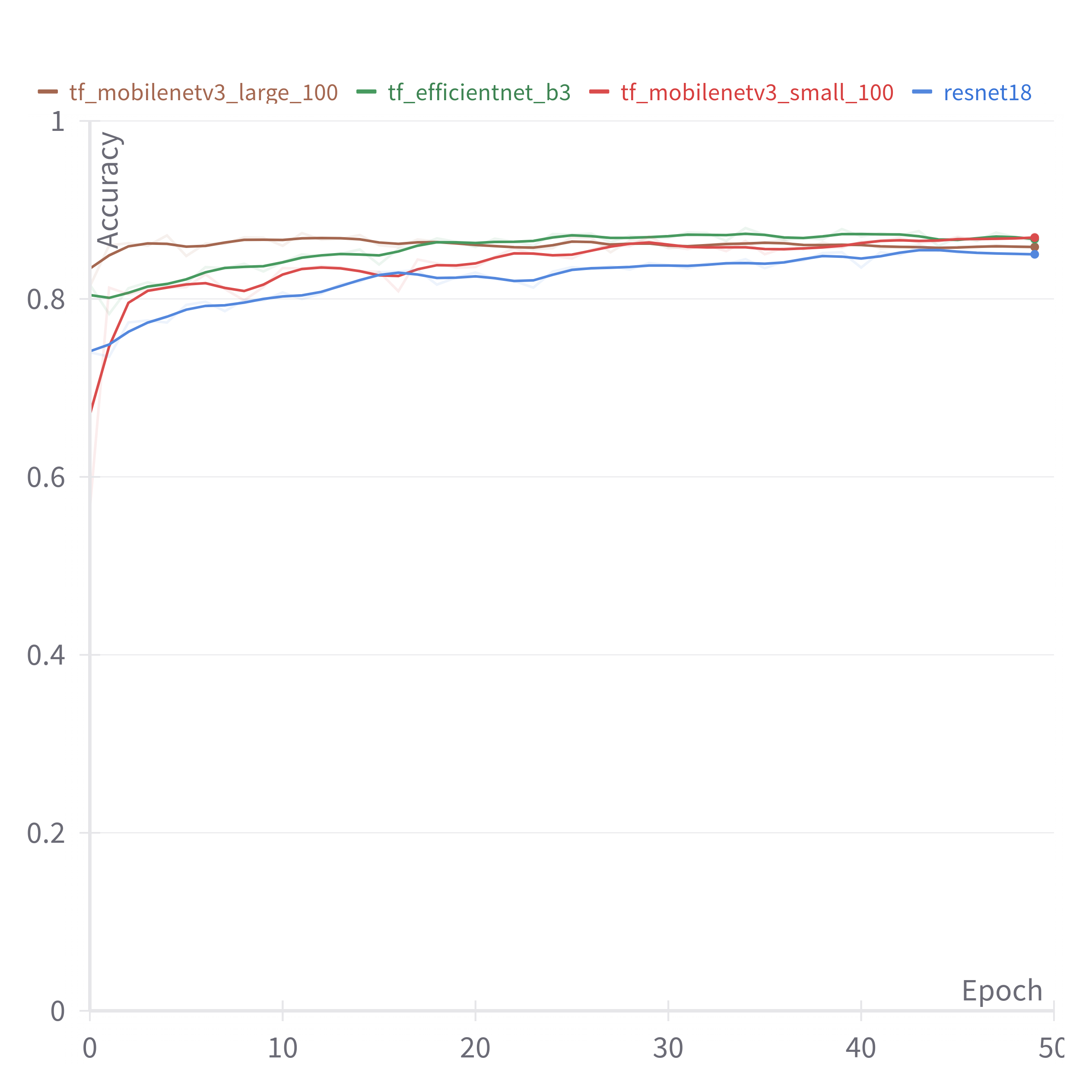}
\caption{\textbf{Validation accuracy performance during training of different models.} We observe that all the models have converged well and show similar performance.}
\label{fig:val_metrics}
\end{figure}

%% file: Sections/pipeline.tex
\input{figures/gradcam}

In this section, we discuss Zeitview's currently deployed turbine crack detection pipeline. The pipeline is split into 3 components: \texttt{image acquisition}, \texttt{inference}, and \texttt{post-processing}. The pipeline is entirely automated until the last stage which requires a human-in-the-loop for final review. A visualization of the end-to-end pipeline is provided in Figure~\ref{fig:pipeline}.

\subsection{Image Acquisition}

For each turbine, we utilize DJI drones to fly a pre-planned flight path which performs a pass of each of a turbine's three blades on all four sides: suction side, pressure side, trailing edge, and leading edge. We utilize a gimbal to orient the camera in such a way that it is orthogonal to the surface of the blade along the entire pass. For each pass, the drone stops to capture an image of the blade at waypoints spaced approximately three meters apart. This means that for a typical 50 meter blade, we capture 18 images per pass, generating a total set of 216 images per turbine.

\subsection{Inference}

After image acquisition, the pilot then uploads the imagery to our cloud infrastructure. Images are then tiled into $1024 \times 1024$ patches.

Once the images have been tiled, we run our classifier in batched processing form and filter out low confidence predictions with a threshold, $\tau=0.5$. Because the images are tiled we have an approximate location on the blade which contains potential cracks. We further localize the cracks into region proposals by computing the gradient attributions w.r.t the crack output class using the GradCAM method~\cite{selvaraju2017grad}.

\input{figures/pipeline}

\subsection{Post-Processing}

GradCAM attribution heatmaps are typically noisy and require additional post-processing to create an accurate localization of detected cracks. Furthermore, storage of heatmaps in the cloud is costly, therefore post-processing to convert region proposals into polygons is necessary. We perform the following steps to convert the attribution heatmaps to localized polygons and provide visual examples in Figure~\ref{fig:gradcam}. 

\begin{description}
    \item[Normalizing] We normalize the attributions within a tile using the min and max statistics.
    \item[Clipping] We remove attribution outliers by clipping the normalized heatmap to the percentile range $[0, 98]$.
    \item[Contour Detection] We utilize contour detection to segment regions with high attribution. We find best results by retrieving only the extreme outer contours and compressing to only obtain the endpoints. We find this as a simple, yet effective method for localizing the cracks within a tile's heatmap.
    \item[Polygonization] Given the raster-level heatmaps outlines, we then convert them into N-point polygons.
    \item[Segmentation] Finally, we utilize our U-Net\cite{ronneberger2015u} based turbine outline segmentation model masks to filter polygons with less than 50\% overlap in pixels with the blade. This model is pretrained for multiclass turbine part segmentation, however, we convert the predictions to binary foreground/background masks for this use case.
\end{description}

\subsection{Human Analysis and Review}

We treat the attribution polygons as region proposals where high severity cracks may exist. From here we incorporate a human-in-the-loop system to refine proposed defects into further severity types and tighter localizations. This is also where any potential false positives are removed. We believe that with further research, we can remove the human from the loop entirely, making the system entirely self-sufficient and scale the number of inspections further.

%% file: figures/gradcam.tex
\begin{figure*}[t]
    \centering
    \includegraphics[width=0.32\textwidth]{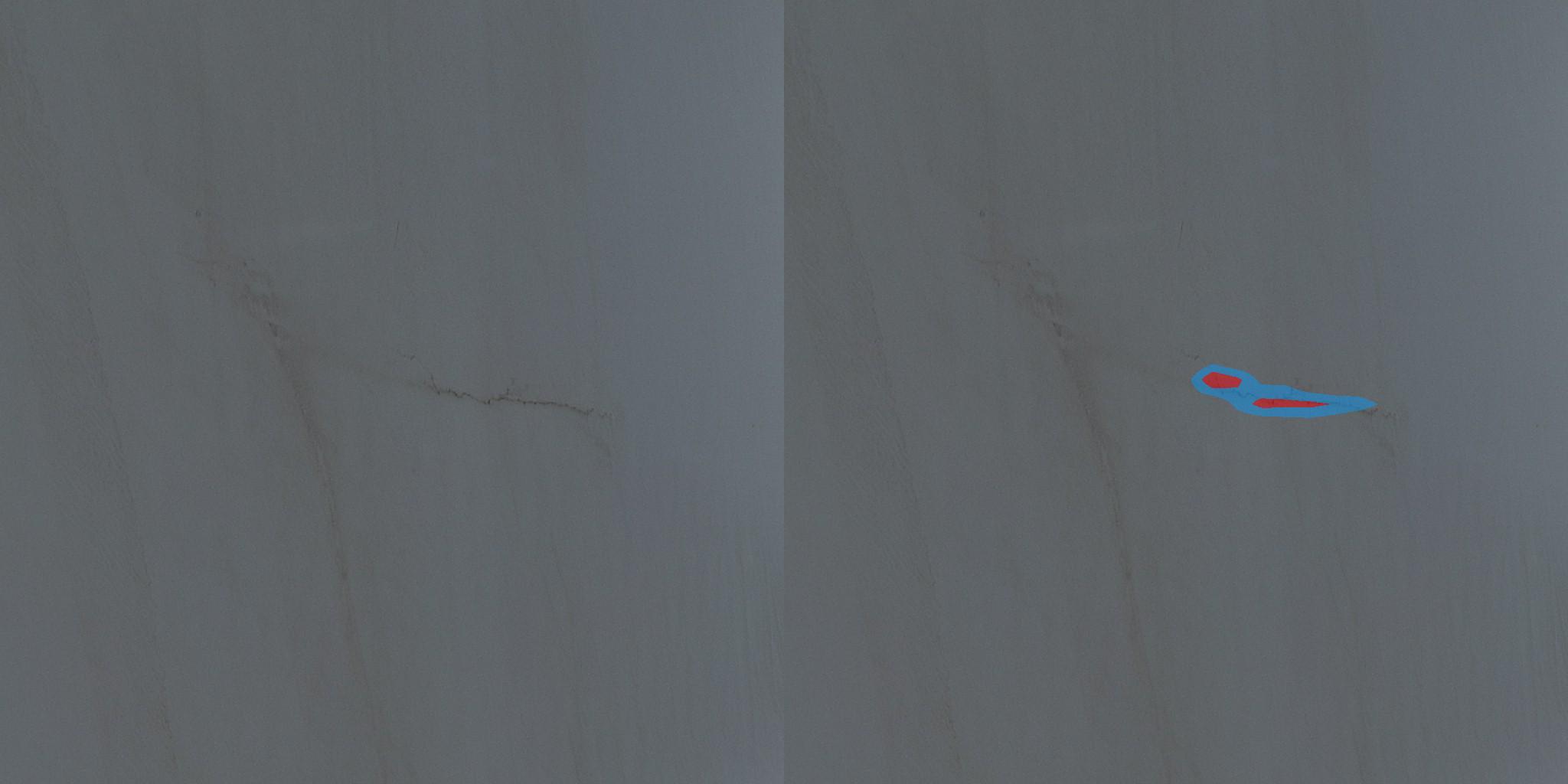}
    \includegraphics[width=0.32\textwidth]{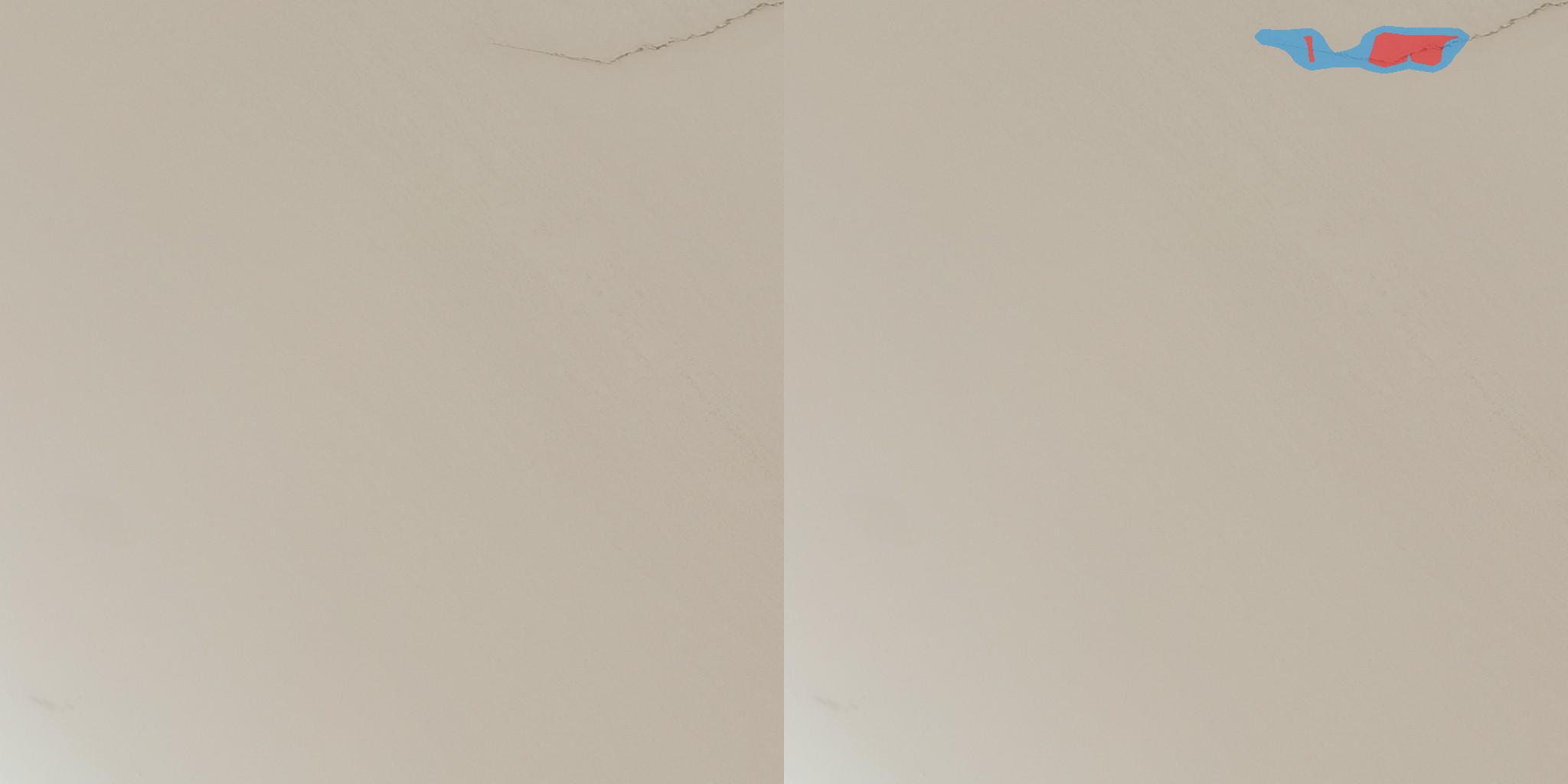}
    \includegraphics[width=0.32\textwidth]{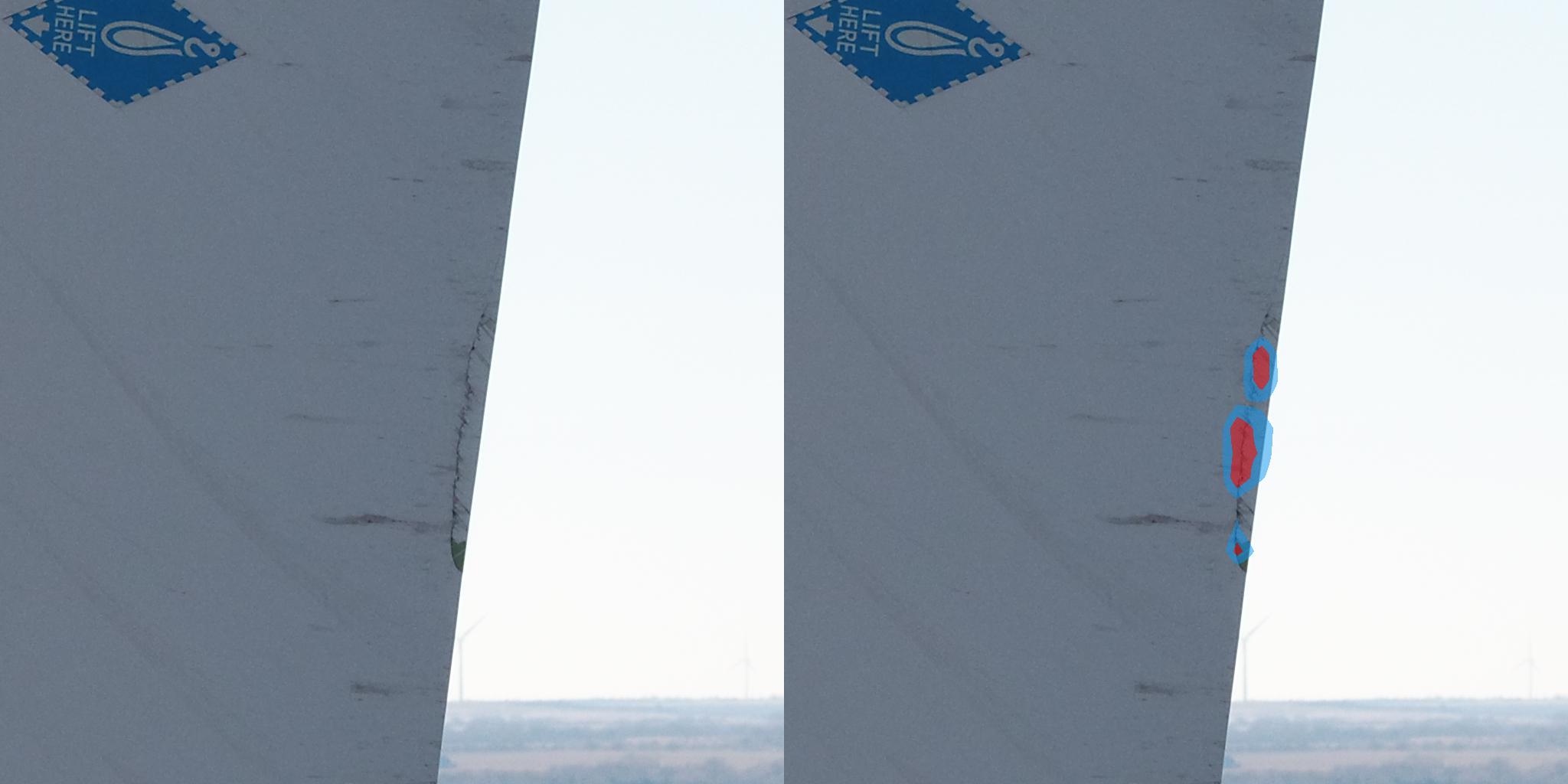} \\
    \includegraphics[width=0.32\textwidth]{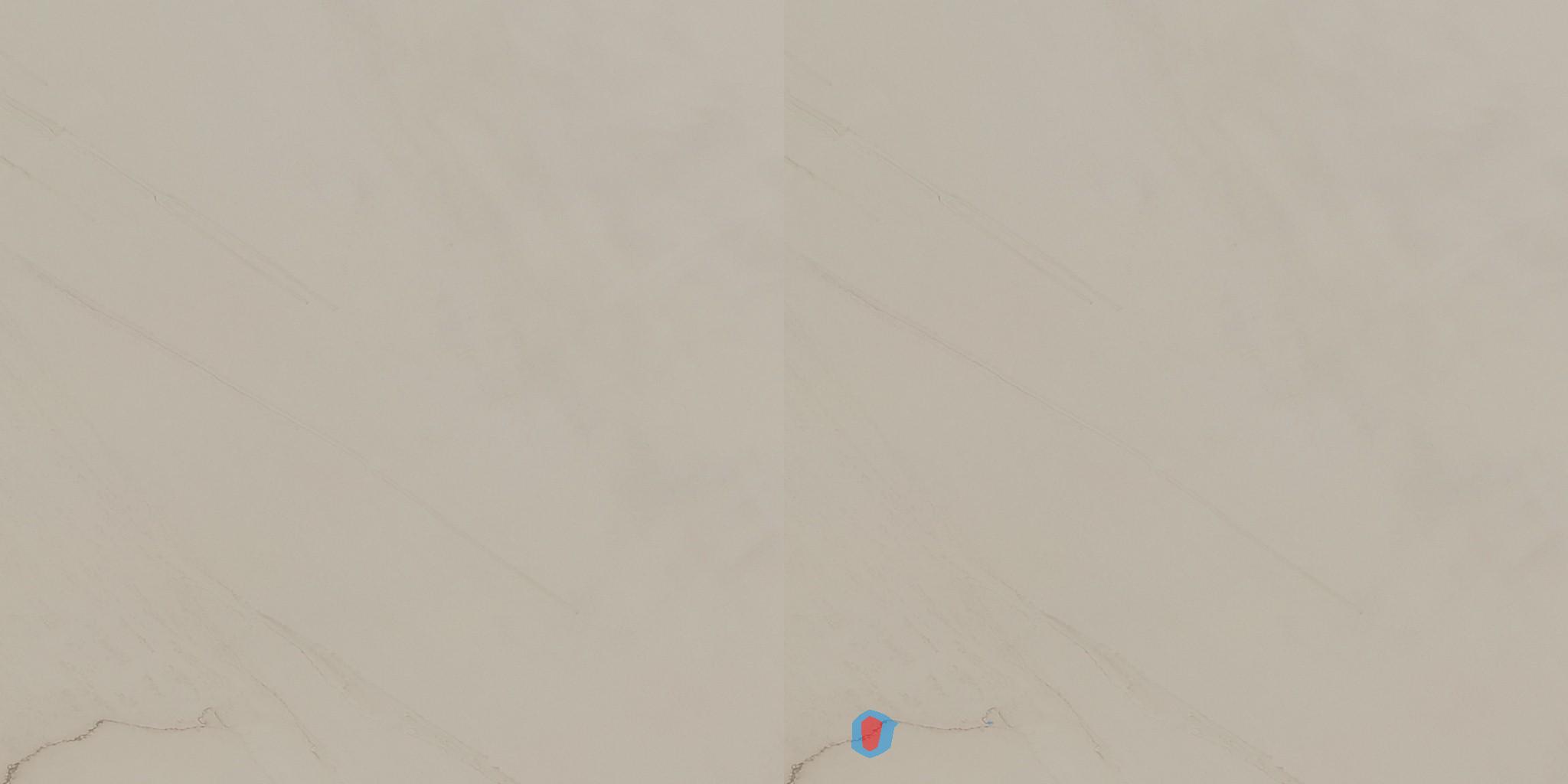}
    \includegraphics[width=0.32\textwidth]{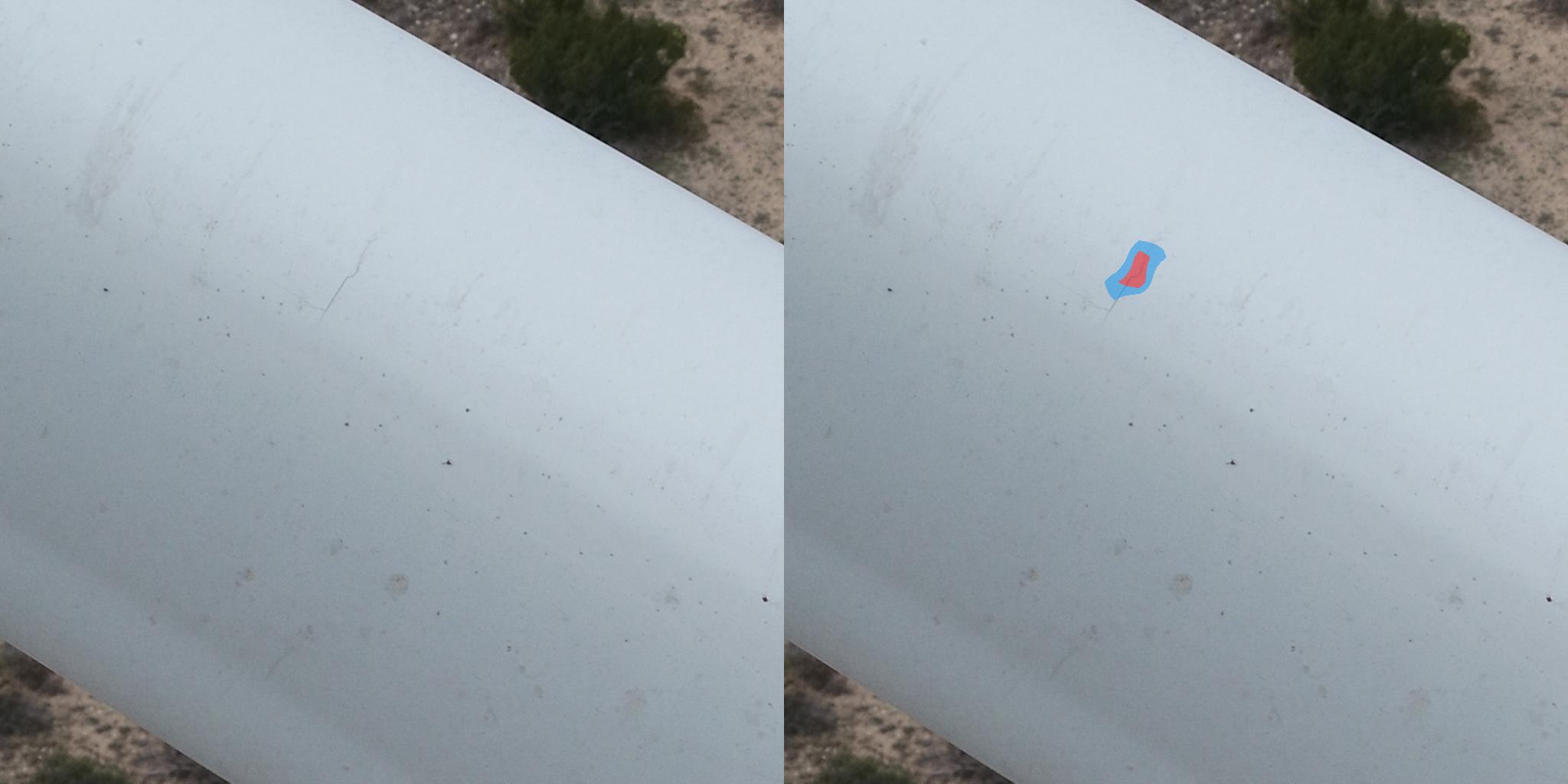}
    \includegraphics[width=0.32\textwidth]{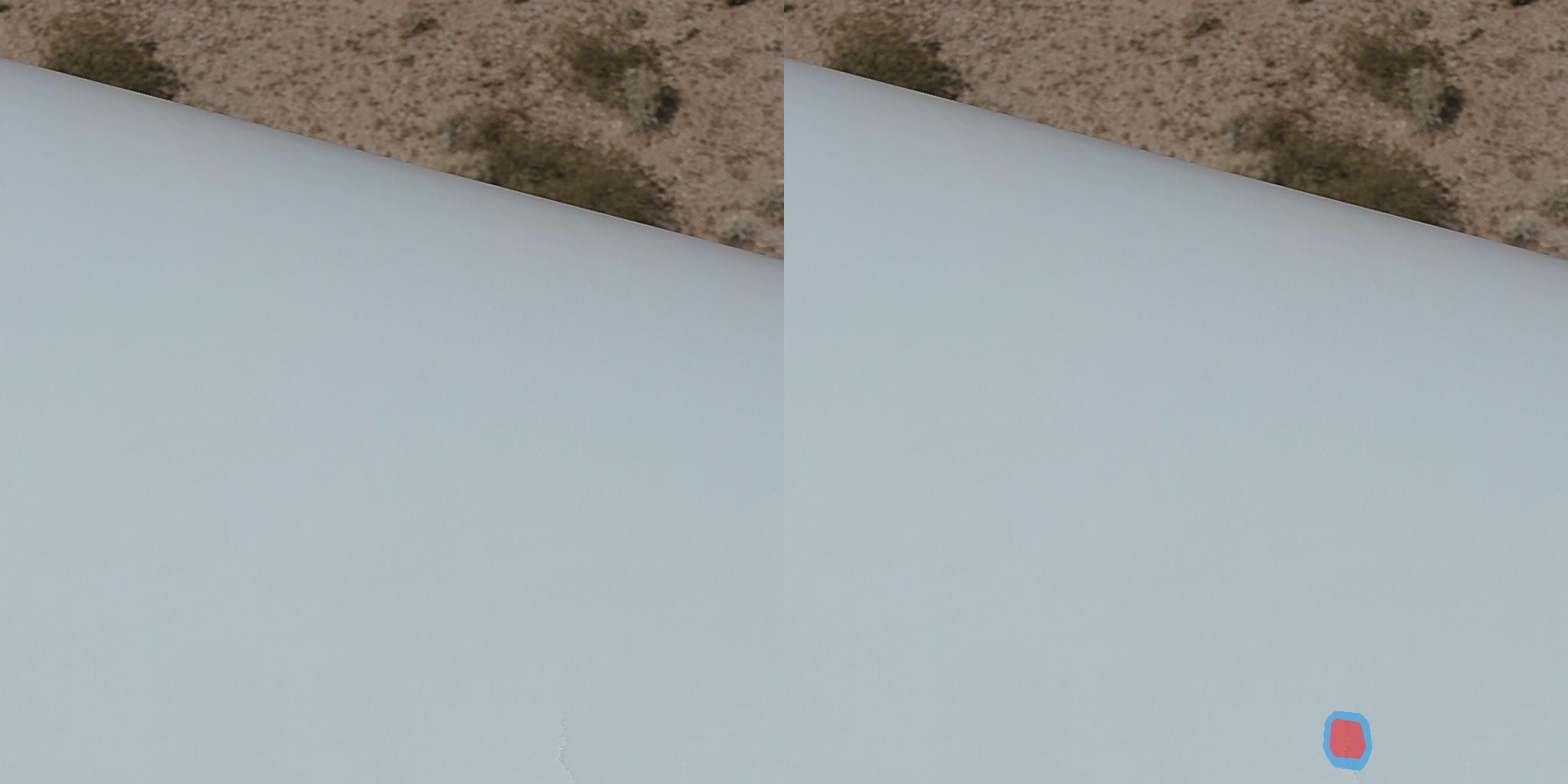} \\
    \includegraphics[width=0.32\textwidth]{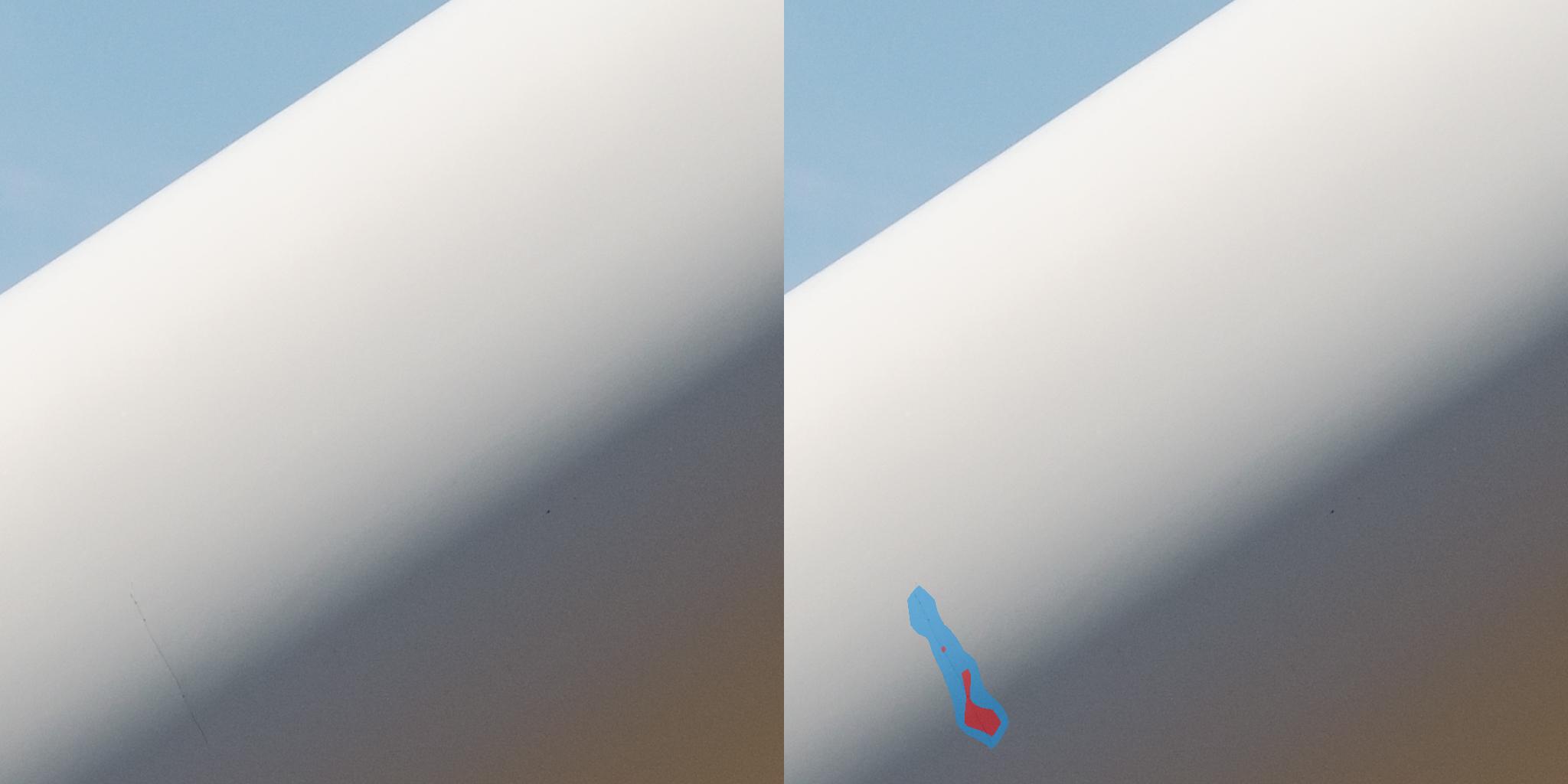}
    \includegraphics[width=0.32\textwidth]{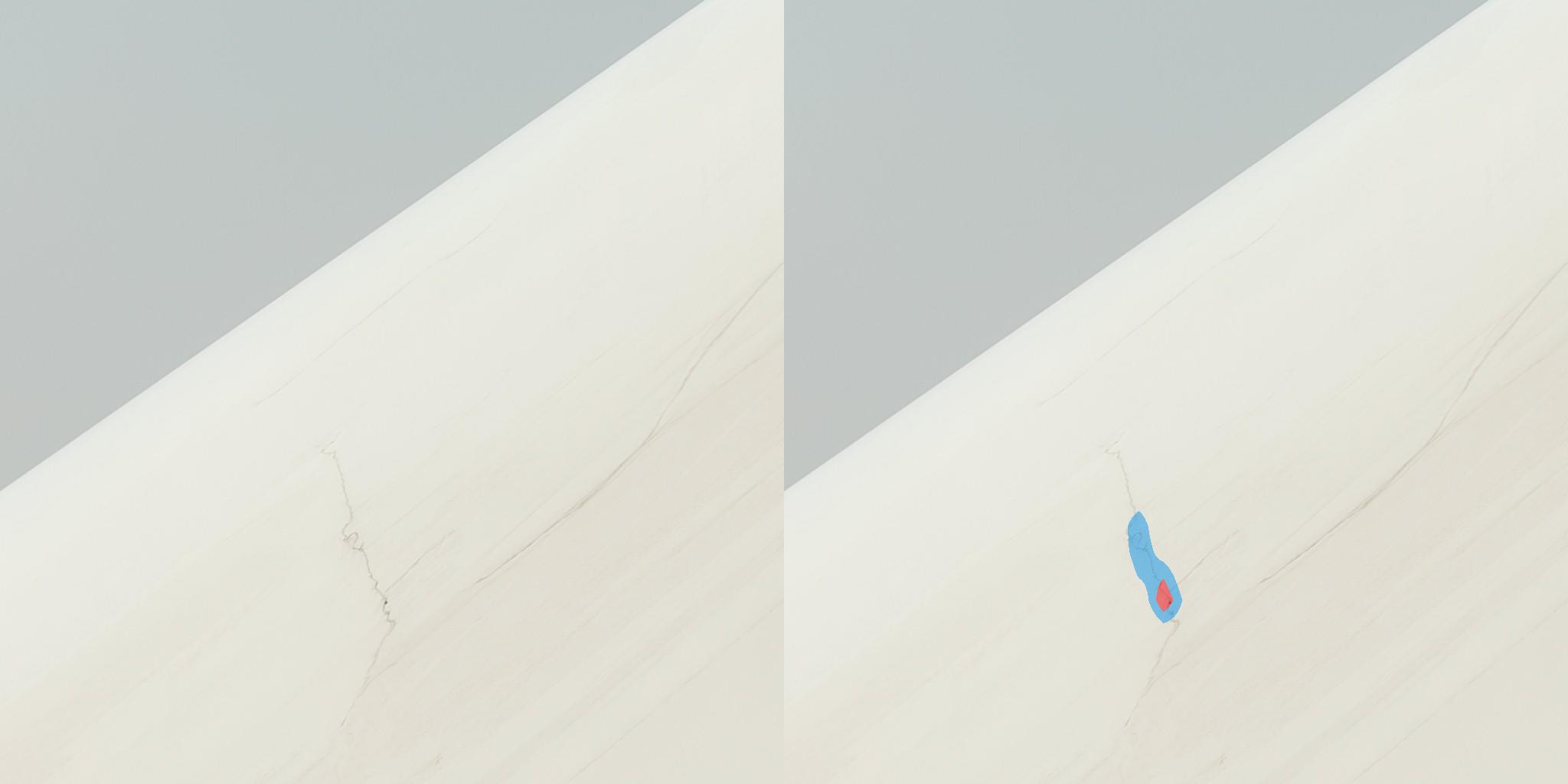}
    \includegraphics[width=0.32\textwidth]{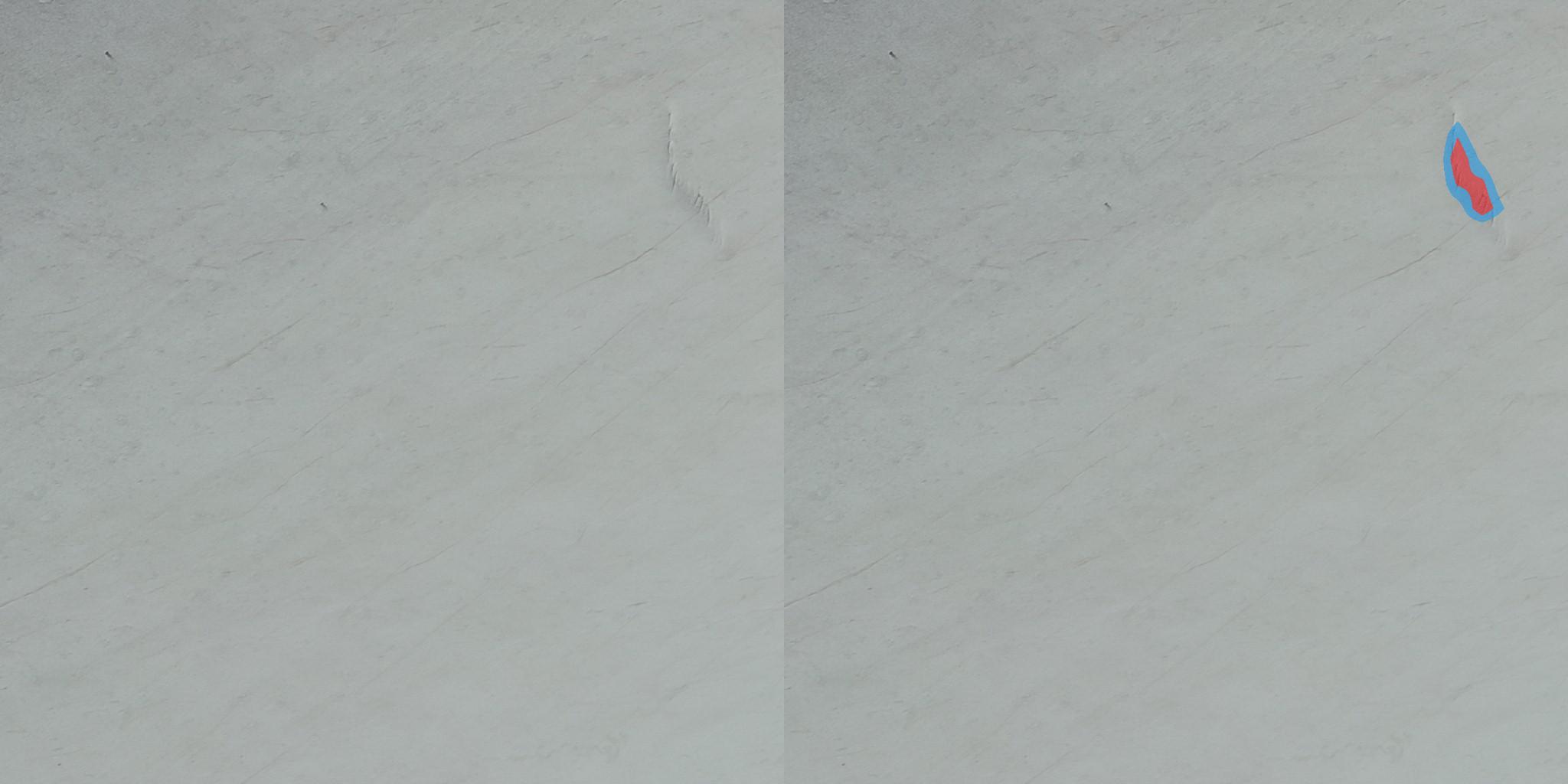} \\
    \includegraphics[width=0.32\textwidth]{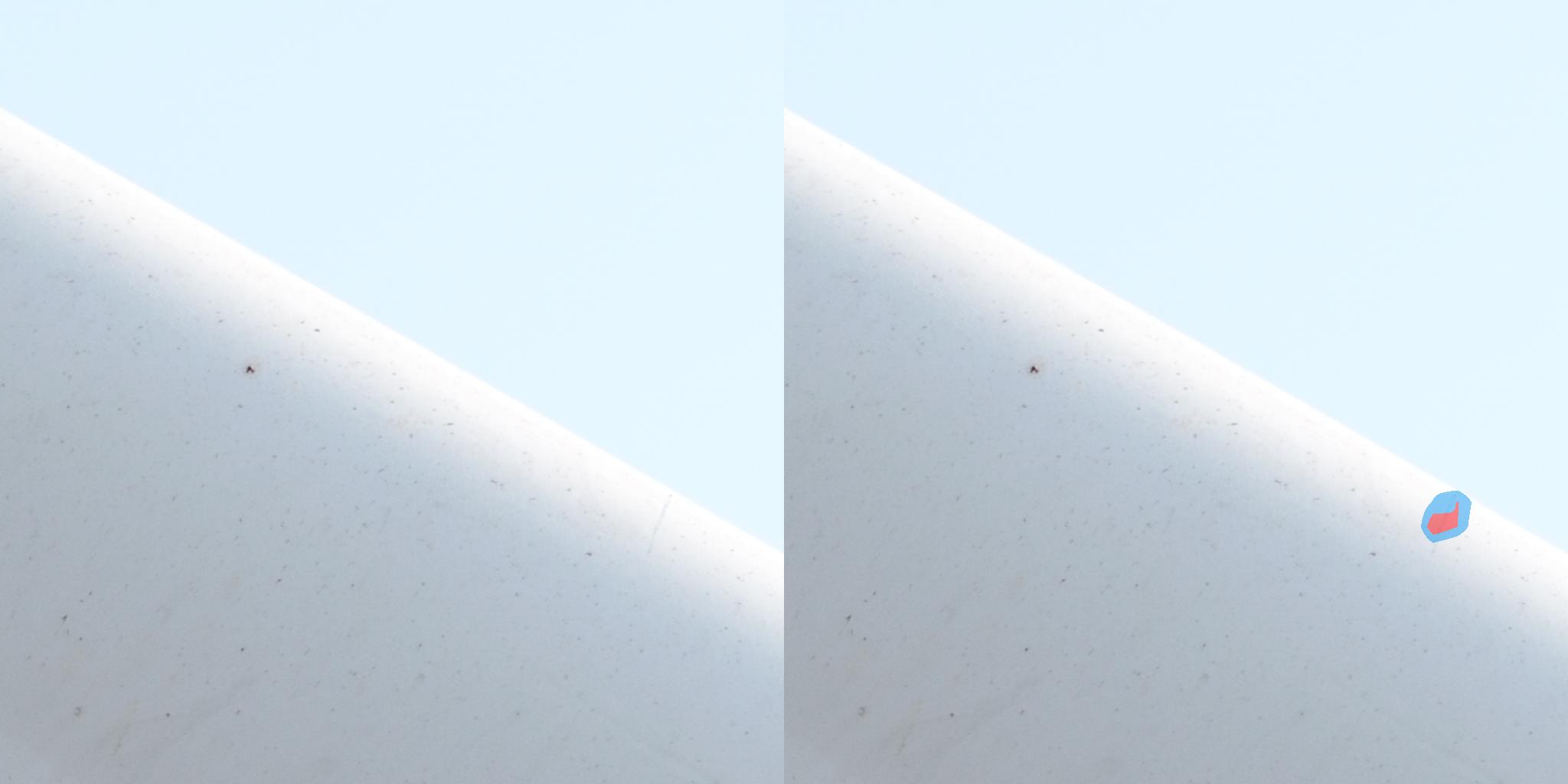}
    \includegraphics[width=0.32\textwidth]{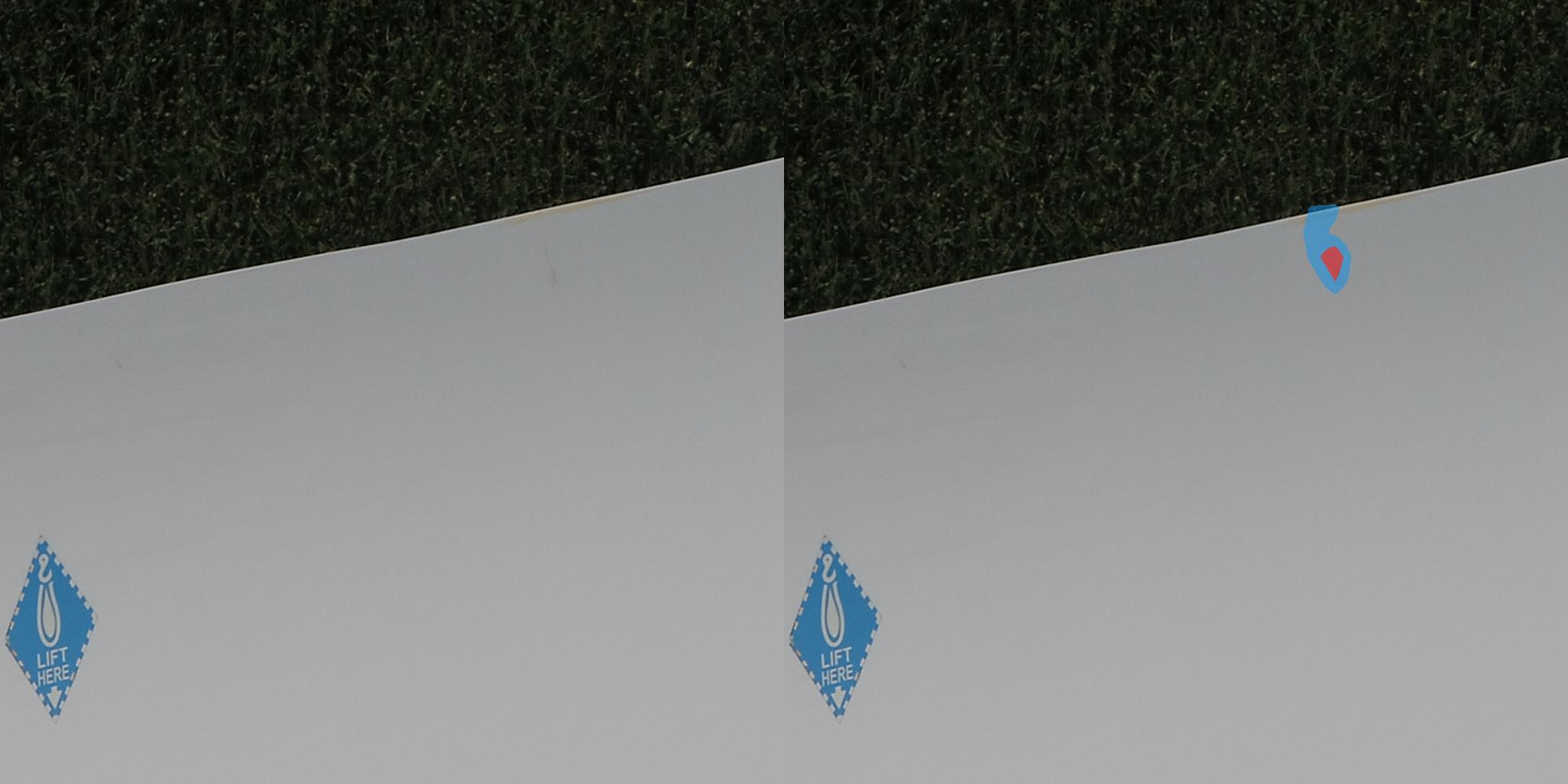}
    \includegraphics[width=0.32\textwidth]{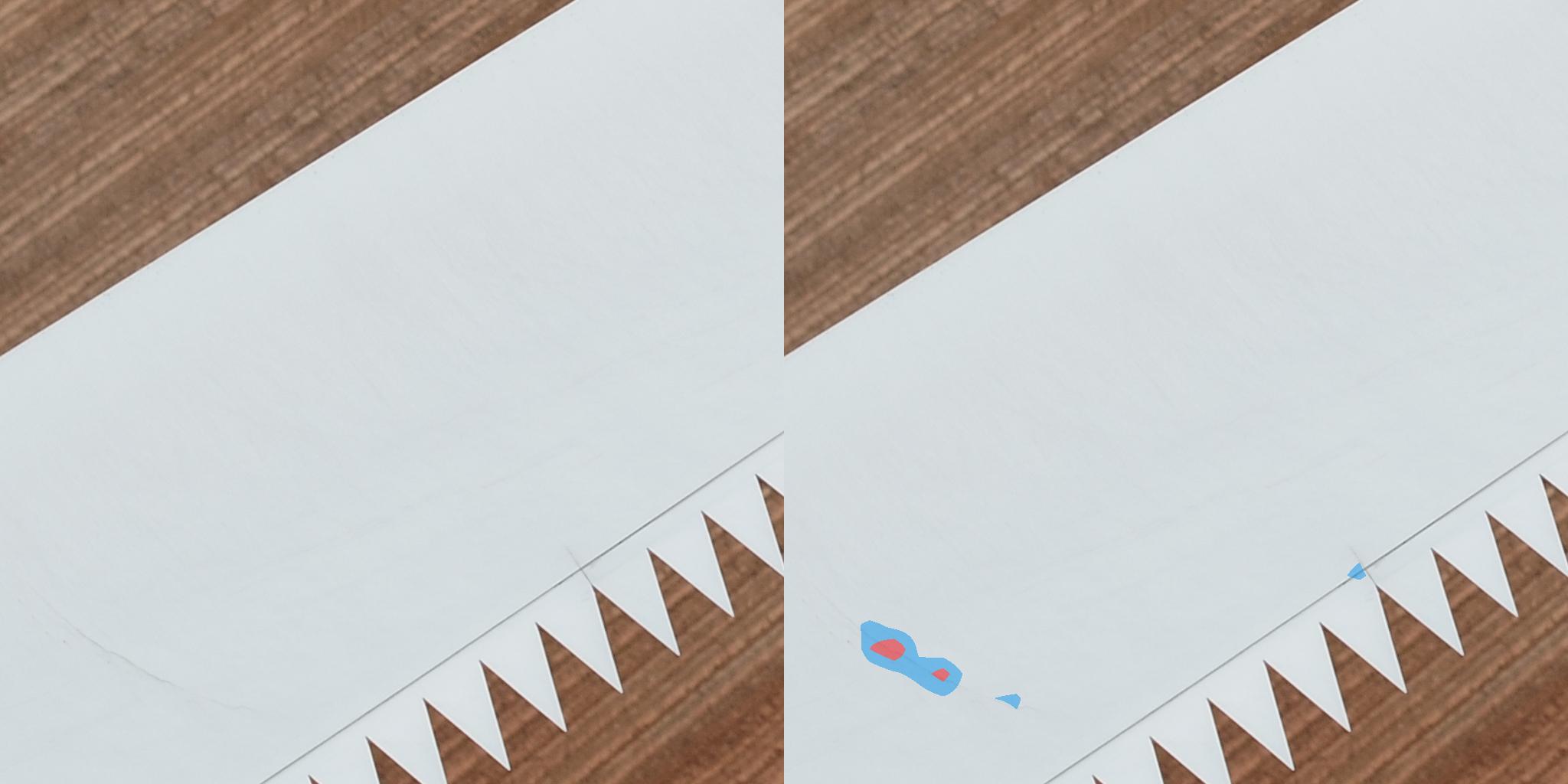}
    \caption{\textbf{Sample Crack Detection Region Proposals using the GradCAM method~\cite{selvaraju2017grad}}. Our pipeline provides region proposals to a human analyst who then reviews and further diagnoses barely-visible surface hairline cracks for severity categorization and repair recommendations. We first compute the GradCAM attributions w.r.t the crack detection class model output. We then post-process the attribution heatmap using the \texttt{normalize} and \texttt{polygonize} operations.}
    \label{fig:gradcam}
\end{figure*}

%% file: figures/pipeline.tex
\begin{figure*}[ht!]
\centering
\includegraphics[width=0.98\linewidth]{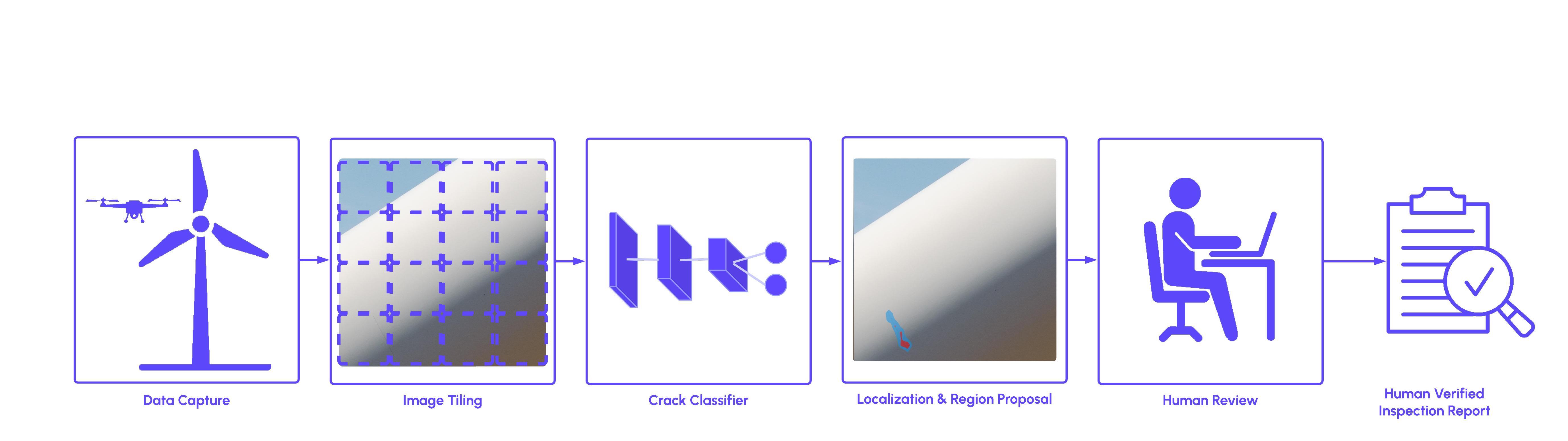}
\caption{\textbf{Our turbine crack detection pipeline diagram.} Our pipeline consists of automated acquisition of overlapping imagery of each turbine blade using DJI UAVs and our flight planning algorithms. Each image is then decomposed into 1024 x 1024 tileswhich we then run our classifier on. Next, we compute and postprocess the gradient attribution heatmap on the high confidence predicted tiles. Our analysts are then shown these proposal regions containing cracks for review and in-depth analysis for inspection report generation.}
\label{fig:pipeline}
\end{figure*}

%% file: Sections/conclusion.tex
In this paper we have proposed a comprehensive, barely-visible, high-severity hairline crack detection dataset for wind turbine inspections, which is significantly more diverse and rich in quality than prior literature. We have also detailed our tiled classification-based approach to hairline crack detection with pseudo region proposals using GradCAM attributions and showed how our system is designed in an end-to-end fashion. We have developed our system with the ability to scale to the vastly increasing number of wind turbines being constructed to generate enough renewable energy to support the world population and reduce reliance on fossil fuels.

We have chosen model architectures which have the ability to be easily ported to drone hardware, allowing inspections to be performed in real time. We hope to move towards this process in future iterations of this research.

Furthermore, we note there has been little research into comparing the performance of infrared vs RGB crack detection. While we have collected a sample dataset for this task, we leave the deeper technical evaluation of this for future work.

Barely-visible surface cracks prove to be a challenging and unsolved problem and can be some of the most severely damaging defects as the turbine is typically only inspected on a yearly basis, giving more than enough time for these minimal cracks to grow and cause catastrophic damage. We hope to further refine our models to improve performance on even the most minute cracks to prevent such disasters from happening. Lastly, we hope this research sheds light to the importance of wind turbine inspection and maintenance to wind energy sustainability.